\documentclass[10pt,twocolumn,letterpaper]{article}

\usepackage[pagenumbers]{cvpr} 

\usepackage{graphicx}
\usepackage{amsmath}
\usepackage{amssymb}
\usepackage{booktabs}
\usepackage{subcaption}
\usepackage{fancyref}

\newcommand{\method}{SUDS}

\usepackage[pagebackref,breaklinks,colorlinks]{hyperref}

\usepackage[capitalize]{cleveref}
\crefname{section}{Sec.}{Secs.}
\Crefname{section}{Section}{Sections}
\Crefname{table}{Table}{Tables}
\crefname{table}{Tab.}{Tabs.}

\begin{document}

\title{SUDS: Scalable Urban Dynamic Scenes}

\author{Haithem Turki\textsuperscript{1*} \qquad Jason Y. Zhang\textsuperscript{1} \qquad Francesco Ferroni\textsuperscript{2} \qquad Deva Ramanan\textsuperscript{1}
\\
\textsuperscript{1}Carnegie Mellon University \qquad \textsuperscript{2}Argo AI } 
\maketitle

\let\thefootnote\relax\footnote{*Work done as an intern at Argo AI.}

\begin{abstract}
   We extend neural radiance fields (NeRFs) to dynamic large-scale urban scenes. Prior work tends to reconstruct single video clips of short durations (up to 10 seconds). Two reasons are that such methods (a) tend to scale linearly with the number of moving objects and input videos because a separate model is built for each and (b) tend to require supervision via 3D bounding boxes and panoptic labels, obtained manually or via category-specific models. As a step towards truly open-world reconstructions of dynamic cities, we introduce two key innovations: (a) we factorize the scene into three separate hash table data structures to efficiently encode static, dynamic, and far-field radiance fields, and (b) we make use of unlabeled target signals consisting of RGB images, sparse LiDAR, off-the-shelf self-supervised 2D descriptors, and most importantly, 2D optical flow.
   Operationalizing such inputs via photometric, geometric, and feature-metric reconstruction losses enables \method\ to decompose dynamic scenes into the static background, individual objects, and their motions. When combined with our multi-branch table representation, such reconstructions can be scaled to tens of thousands of objects across 1.2 million frames from 1700 videos spanning geospatial footprints of hundreds of kilometers, (to our knowledge) the largest dynamic NeRF built to date.
   We present qualitative initial results on a variety of tasks enabled by our representations, including novel-view synthesis of dynamic urban scenes, unsupervised 3D instance segmentation, and unsupervised 3D cuboid detection. To compare to prior work, we also evaluate on KITTI and Virtual KITTI 2, surpassing state-of-the-art methods that rely on ground truth 3D bounding box annotations while being 10x quicker to train.
\end{abstract}

\section{Introduction}

\begin{figure}
     \centering
         \includegraphics[width=\linewidth, clip=true, trim = 0mm 0mm 0mm 0mm]{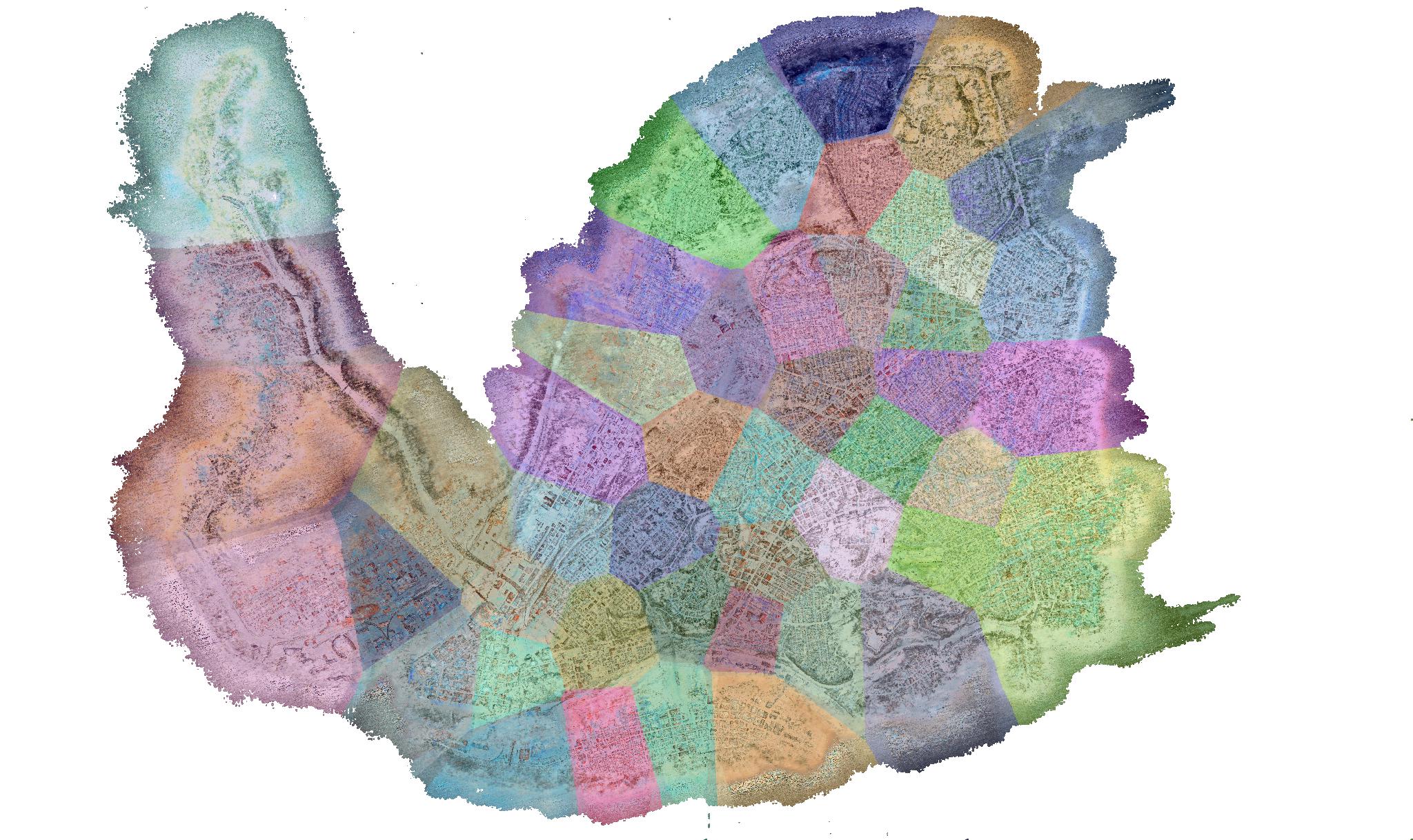}\\
         \includegraphics[width=\linewidth, clip=true, trim = 0mm 2mm 0mm 0mm]{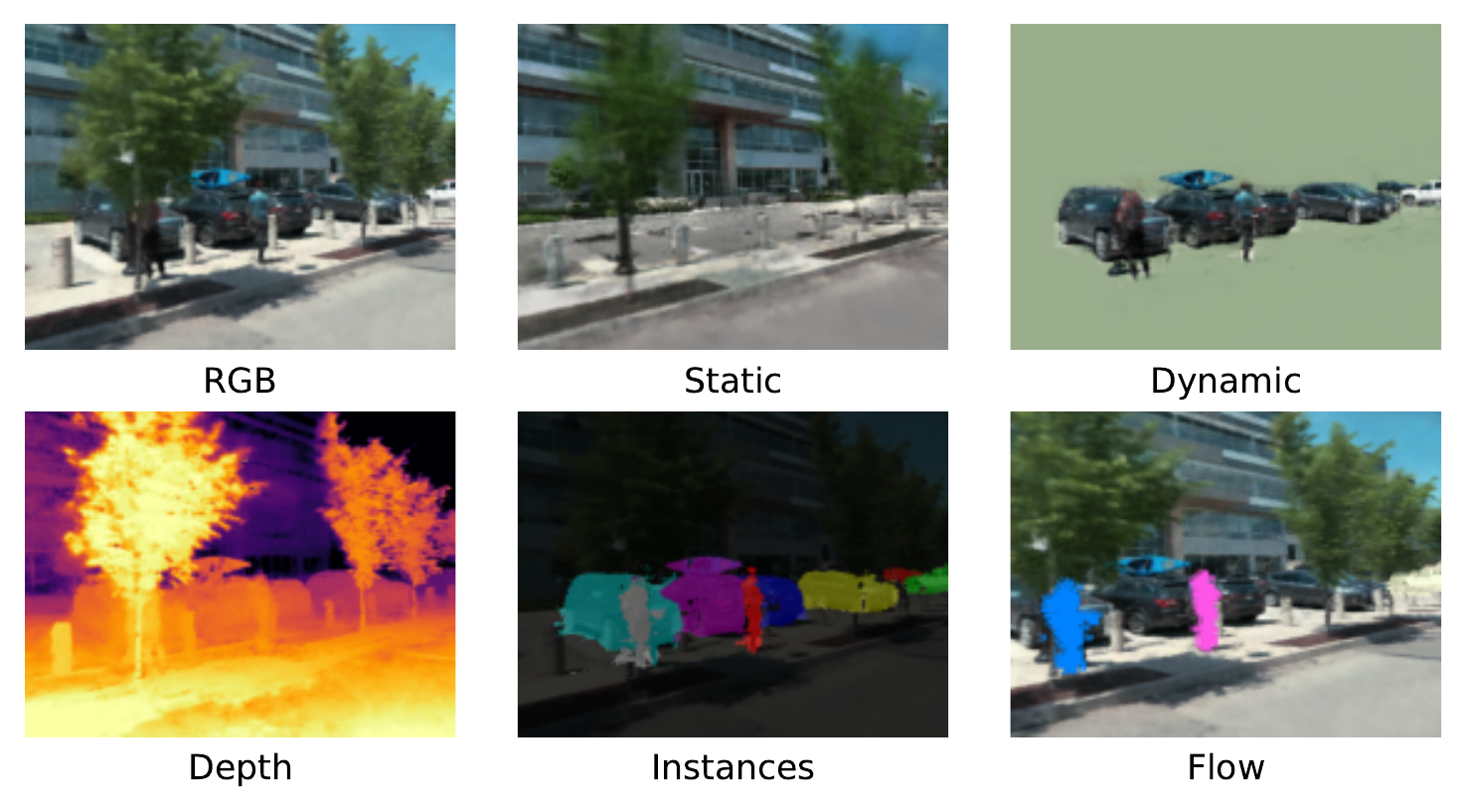}
\vspace*{-6mm}
   \caption{{\bf SUDS.} We scale neural reconstructions to city scale by dividing the area into multiple cells and training hash table representations for each. We show our full city-scale reconstruction {\bf above} and the derived representations {\bf below.} Unlike prior methods, our approach handles dynamism across multiple videos, disentangling dynamic objects from static background and modeling shadow effects. We use unlabeled inputs to learn scene flow and semantic predictions, enabling category- and object-level scene manipulation.}
\vspace*{-6mm}
\label{fig:teaser}
\end{figure}

Scalable geometric reconstructions of cities have transformed our daily lives, with tools such as Google Maps and Streetview~\cite{anguelov2010google} becoming fundamental to how we navigate and interact with our environments. A watershed moment in the development of such technology was the ability to scale structure-from-motion (SfM) algorithms to city-scale footprints~\cite{agarwal2011building}.
Since then, the advent of Neural Radiance Fields (NeRFs)~\cite{mildenhall2020nerf} has transformed this domain by allowing for photorealistic interaction with a reconstructed scene via view synthesis.

Recent works have attempted to scale such representations to neighborhood-scale reconstructions for virtual drive-throughs~\cite{tancik2022blocknerf} and photorealistic fly-throughs~\cite{Turki_2022_CVPR}. However, these maps remain static and frozen in time. This makes capturing bustling human environments---complete with moving vehicles, pedestrians, and objects---impossible, limiting the usefulness of the representation.

{\bf Challenges.} One possible solution is a dynamic NeRF that conditions on time or warps a canonical space with a time-dependent deformation~\cite{park2021nerfies}. However, reconstructing dynamic scenes is notoriously challenging because the problem is inherently under-constrained, particularly when input data is constrained to limited viewpoints, as is typical from egocentric video capture~\cite{gao2022dynamic}. One attractive solution is to scale up reconstructions to {\em many} videos, perhaps collected at different days (e.g., by an autonomous vehicle fleet). However, this creates additional challenges in jointly modeling fixed geometry that holds for all time (such as buildings), geometry that is locally static but {\em transient} across the videos (such as a parked car), and geometry that is truly dynamic (such as a moving person).

{\bf \method.} In this paper, we propose \method: Scalable Urban Dynamic Scenes, a 4D representation that targets both \textit{scale} and \textit{dynamism}. Our key insight is twofold; (1) \method\ makes use of a rich suite of informative but freely available input signals, such as LiDAR depth measurements and optical flow. Other dynamic scene representations~\cite{KunduCVPR2022PNF,Ost_2021_CVPR} require supervised inputs such as panoptic segmentation labels or bounding boxes, which are difficult to acquire with high accuracy for our in-the-wild captures. (2) \method\ decomposes the world into 3 components: a \textit{static} branch that models stationary topography that is consistent across videos, a \textit{dynamic} branch that handles both transient (e.g., parked cars) and truly dynamic objects (e.g., pedestrians), and an \textit{environment map} that handles far-field objects and sky. We model each branch using a multi-resolution hash table with scene partitioning, allowing \method\ to scale to an entire city spanning over 100 $km^2$. 

{\bf Contributions.} 
We make the following contributions: (1) to our knowledge, we build the first large-scale dynamic NeRF, (2) we introduce a scalable three-branch hash table representation for 4D reconstruction, (3) we present state-of-the-art reconstruction on 3 different datasets. Finally, (4) we showcase a variety of downstream tasks enabled by our representation, including free-viewpoint synthesis, 3D scene flow estimation, and even unsupervised instance segmentation and 3D cuboid detection.

\begin{figure*}[t!]
  \vspace*{-1mm}
  \centering
  \begin{subfigure}[ht]{0.35\textwidth}
    \centering
    \includegraphics[width=0.61\textwidth]{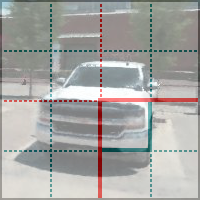}
    \vspace*{0.3mm}
    \caption{\textbf{Voxel Lookup}}
  \end{subfigure}
  \hspace*{-6mm}
  \begin{subfigure}[ht]{0.12\textwidth}
    \vspace*{1mm}
    \begin{subfigure}{0.9\textwidth}
      \centering
      \includegraphics[width=0.5\textwidth]{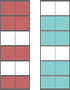}
      \caption*{$\text{static\_hash}(\textbf{v}_{l,s})$}
    \vspace*{3.3mm}
    \end{subfigure}
    \begin{subfigure}{0.9\textwidth}
      \centering
      \includegraphics[width=0.5\textwidth]{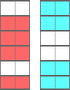}
      \caption*{$\text{dynamic\_hash}(\textbf{v}_{l,d})$}
    \vspace*{0.4mm}
    \end{subfigure}
    \caption{\textbf{Indexing}}
  \end{subfigure}
  \hspace*{-1.2mm}
  \begin{subfigure}[ht]{0.1\textwidth}
    \vspace*{1mm}
    \begin{subfigure}{0.5\textwidth}
      \centering
      \footnotesize{$\quad\quad\quad\quad\textbf{d}$$\rightarrow$}
      \footnotesize{$\textbf{ }A_{vid}\mathcal{F}(t)$$\rightarrow$}
      \vspace*{11mm}
    \end{subfigure}
    \begin{subfigure}{0.5\textwidth}
      \centering
      \footnotesize{$\quad\quad\quad\quad\textbf{d}$$\rightarrow$}
    \end{subfigure}
  \end{subfigure}
  \hspace*{-7mm}
  \begin{subfigure}[ht]{0.17\textwidth}
  \vspace*{1.5mm}
    \begin{subfigure}{\textwidth}
      \centering
      \includegraphics[width=0.72\textwidth]{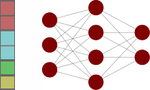}
    \caption*{$(\textbf{c}_s, \sigma_s, \phi_s)$}
    \vspace*{3mm}
    \end{subfigure}
    \begin{subfigure}{\textwidth}
      \centering
      \includegraphics[width=0.72\textwidth]{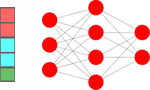}
    \caption*{$(\textbf{c}_d, \sigma_d, \phi_d, \rho_d, s_{\textbf{t}-1}, s_{\textbf{t}+1})$}
    \vspace*{-1mm}
    \end{subfigure}
    \caption{\textbf{MLP Evaluation}}
    \label{fig:flat}
  \end{subfigure}
  \begin{subfigure}[ht]{0.2\textwidth}
    \centering
    $(\textbf{c}, \sigma, \phi, s_{\textbf{t}-1}, s_{\textbf{t}+1})$
    \caption{\textbf{Output Blending}}
  \end{subfigure}
\caption{{\bf Model Architecture.} (a) For a given input coordinate, we find the surrounding voxels at $L$ resolution levels for both the static and dynamic branches (far-field branch omitted for clarity). (b) We assign indices to their corners by hashing based on position in the static branch and position, time, and video id in the dynamic branch. We look up the feature vectors corresponding to the corners and interpolate according to the relative position of the input coordinate within the voxel. (c) We concatenate the result of each level, along with auxiliary inputs such as viewing direction, and pass the resulting vector into an MLP to obtain per-branch color, density, and feature logits along with scene flow and the shadow ratio. (d) We blend scolor, opacity, and feature logits as the weighted sum of the branches.}
  \label{fig:architecture}
  \vspace*{-4mm}
\end{figure*}

\section{Related Work}

The original Neural Radiance Fields (NeRF) paper~\cite{mildenhall2020nerf} inspired a wide body of follow-up work based on the original approach. Below, we describe a non-exhaustive list of such approaches along axes relevant to our work.

\textbf{Scale.} The original NeRF operated with bounded scenes. NeRF++~\cite{zhang2020npp} and mip-NeRF 360~\cite{barron2022mipnerf360} use non-linear scene parameterization to model unbounded scenes. However, scaling up the size of the scene with a fixed size MLP leads to blurry details and training instability while the cost of naively increasing the size of the MLP quickly becomes intractable. BungeeNeRF~\cite{xiangli2022bungeenerf} introduced a coarse-to-fine approach that progressively adds more capacity to the network representation. 
Block-NeRF~\cite{tancik2022blocknerf} and Mega-NeRF~\cite{Turki_2022_CVPR} partition the scene spatially and train separate NeRFs for each partition. To model appearance variation, they incorporate per-image embeddings like NeRF-W~\cite{martinbrualla2020nerfw}. Our approach similarly partitions the scene into sub-NeRFs, making use of depth to improve partition efficiency and scaling over an area 200x larger than Block-NeRF's Alamo Square Dataset. Both of these methods work only on static scenes.

\textbf{Dynamics.} Neural 3D Video Synthesis~\cite{li2021neural} and Space-time Neural Irradiance Fields~\cite{xian2021space} add time as an input to handle dynamic scenes. Similar to our work, NSFF~\cite{li2020neural}, NeRFlow~\cite{du2021nerflow}, and DyNeRF~\cite{Gao-ICCV-DynNeRF} incorporate 2D optical flow input and warping-based regularization losses to enforce plausible transitions between observed frames. Multiple methods ~\cite{park2021nerfies, pumarola2020d, tretschk2021nonrigid, park2021hypernerf} instead disentangle scenes into a canonical template and per-frame deformation field. BANMo~\cite{yang2022banmo} further incorporates deformable shape models and canonical embeddings to train articulated 3D models from multiple videos. 
These methods focus on single-object scenes, and all but \cite{li2021neural} and \cite{yang2022banmo} use single video sequences.

While many of the previous works use segmentation data to factorize dynamic from static objects, D$^2$NeRF~\cite{wu2022d} does this automatically through regularization and explicitly handling shadows. Neural Groundplans~\cite{see3d} uses synthetic data to do this decomposition from a single image.  We borrow some of these ideas and scale beyond synthetic and indoor scenes.

\textbf{Object-centric approaches.} Several approaches~\cite{Niemeyer2020GIRAFFE, Ost_2021_CVPR, zhang2021stnerf, yang2021objectnerf, yu2022unsupervised, yuan2021star} represent scenes as the composition of per-object NeRF models and a background model. NSG~\cite{Ost_2021_CVPR} is most similar to us as it also targets automotive data but cannot handle ego-motion as our approach can. None of these methods target multi-video representations and are fundamentally constrained by the memory required to represent each object, with NSG needing over 1TB of memory to represent a 30 second video in our experience.

\textbf{Semantics.} Follow-up works have explored additional semantic outputs in addition to predicting color. Semantic-NeRF~\cite{Zhi:etal:ICCV2021} adds an extra head to NeRF that predicts extra semantic category logits for any 3D position. Panoptic-NeRF~\cite{fu2022panoptic} and Panoptic Neural Fields~\cite{KunduCVPR2022PNF} extend this to produce panoptic segmentations and the latter uses a similar bounding-box based object and background decomposition as NSG. NeSF~\cite{vora2021nesf} generalizes the notion of a semantic field to unobserved scenes. As these methods are highly reliant on accurate annotations which are difficult to reliably obtain in the wild at our scale, we instead use a similar approach to recent works~\cite{kobayashi2022distilledfeaturefields, tschernezki22neural} that distill the outputs of 2D self-supervised feature descriptors into 3D radiance fields to enable semantic understanding without the use of human labels and extend them to larger dynamic settings.

\textbf{Fast training.} The original NeRF took 1-2 days to train.
Plenoxels~\cite{yu_and_fridovichkeil2021plenoxels} and DVGO~\cite{SunSC22} directly optimize a voxel representation instead of an MLP to train in minutes or even seconds. TensoRF~\cite{Chen2022ECCV} stores its representation as the outer product of low-rank tensors, reducing memory usage. Instant-NGP~\cite{mueller2022instant} takes this further by encoding features in a multi-resolution hash table, allowing training and rendering to happen in real-time. We use these tables as the base block of our three-branch representation and use our own hashing method to support dynamics across multiple videos.

\textbf{Depth.} Depth provides a valuable supervisory signal for learning high-quality geometry. DS-NeRF~\cite{kangle2021dsnerf} and Dense Depth Priors~\cite{roessle2022depthpriorsnerf} incorporate noisy point clouds obtained by structure from motion (SfM) in the loss function during optimization. Urban Radiance Fields~\cite{rematas2022urf} supervises with collected LiDAR data. We also use LiDAR but demonstrate results on dynamic environments.

\section{Approach}

\subsection{Inputs}

Our goal is to learn a global representation that facilitates free-viewpoint rendering, semantic decomposition, and 3D scene flow at arbitrary poses and time steps. Our method takes as input ordered RGB images from $N$ videos (taken at different days with diverse weather and lighting conditions) and their associated camera poses.
Crucially, we make use of additional data as ``free" sources of supervision given contemporary sensor rigs and feature descriptors. Specifically, we use (1) aligned sparse LiDAR depth measurements, (2) 2D self-supervised pixel (DINO~\cite{caron2021emerging}) descriptors to enable semantic manipulation, and (3) 2D optical flow predictions to model scene dynamics. All model inputs are generated without any human labeling or intervention.

\subsection{Representation}
\label{sec:representation}

\textbf{Preliminaries.} We build upon NeRF~\cite{mildenhall2020nerf}, which represents a scene within a continuous volumetric radiance field that captures both geometry and view-dependent appearance. It encodes the scene within the weights of a multilayer perceptron (MLP). At render time, NeRF projects a camera ray $\textbf{r}$ for each image pixel and samples along the ray, querying the MLP at sample position $\textbf{x}_i$ and ray viewing direction $\textbf{d}$ to obtain opacity and color values $\sigma_i$ and $\textbf{c}_i$. It then composites a color prediction $\hat{C}(\textbf{r})$ for the ray using numerical quadrature $\sum_{i=0}^{N-1} T_i (1 - \exp( -\sigma_{i} \delta_{i})) \, \textbf{c}_i$, where $T_i = \exp( -\sum_{j=0}^{i-1} \sigma_j \delta_j)$ and $\delta_i$ is the distance between samples. The training process optimizes the model by sampling batches $R$ of image pixels and minimizing the loss function $\sum_{\textbf{r} \in \mathcal{R}} \big\lVert{C(\textbf{r}) - \hat{C}(\textbf{r})}\big\rVert^2$. NeRF samples rays through a two-stage hierarchical sampling process and uses frequency encoding to capture high-frequency details. We refer the reader to~\cite{mildenhall2020nerf} for more details.

\textbf{Scene composition.} To model large-scale dynamic environments, \method\ factorizes the scene into three branches: (a) a static branch containing non-moving topography consistent across videos, (b) a dynamic branch to disentangle video-specific objects~\cite{Gao-ICCV-DynNeRF, li2020neural, wu2022d}, moving or otherwise, and (c) a far-field environment map to represent far-away objects and the sky, which we found important to separately model in large-scale urban scenes~\cite{zhang2020npp, Turki_2022_CVPR, rematas2022urf}.

However, conventional NeRF training with MLPs is computationally prohibitive at our target scales. Inspired by Instant-NGP~\cite{mueller2022instant}, we implement each branch using multiresolution hash tables of $F$-dimensional feature vectors followed by a small MLP, along with our own hash functions to index across videos.

\textbf{Hash tables (Fig.~\ref{fig:architecture})}. For a given input coordinate $(\textbf{x}, \textbf{d}, \textbf{t}, \textbf{vid})$ denoting the position ${\bf x} \in \mathbb{R}^3$, viewing direction ${\bf d} \in \mathbb{R}^3$, frame index $F \in \{1,...,T\}$, and video id ${\bf vid} \in \{1,...,N\}$, we find the surrounding voxels in each table at $l \in L$ resolution levels, doubling the resolution between levels, which we denote as $\textbf{v}_{l,s}$, $\textbf{v}_{l,d}$, $\textbf{v}_{l,e}$ for the static, dynamic, and far-field. The static branch makes use of 3D spatial voxels $\textbf{v}_{l,s}$, while the dynamic branch makes use of 4D spacetime voxels $\textbf{v}_{l,d}$. Finally, the far-field branch makes use of 3D voxels $\textbf{v}_{l,e}$ (implemented via normalized 3D direction vectors) that index an environment map. Similar to Instant-NGP~\cite{mueller2022instant}, rather than storing features at voxel corners, we compute hash indices $\textbf{i}_{l,s}$ (or $\textbf{i}_{l,d}$ or $\textbf{i}_{l,e}$) for each corner with the following hash functions:
\begin{align}
    \textbf{i}_{l,s} &= \text{static\_hash}({space}(\textbf{v}_{l,s})) \\
    \textbf{i}_{l,d} &= \text{dynamic\_hash}({space}(\textbf{v}_{l,d}), {time}(\textbf{v}_{l,d}), \textbf{vid})\\
    \textbf{i}_{l,e} &= \text{env\_hash}({dir}(\textbf{v}_{l,e}), \textbf{vid})
\end{align}

We linearly interpolate features up to the nearest voxel vertices (but now relying on {\em quad}linear interpolation for the dynamic 4D branch) and rely on gradient averaging to handle hash collisions. Finally, to model the fact that different videos likely contain distinct moving objects and illumination conditions, we add $\textbf{vid}$  as an auxiliary input to the hash, but do {\em not} use it for interpolation (since averaging across distinct movers is unnatural). From this perspective, we leverage hashing to effectively index separate interpolating functions for each video, {\em without} a linear growth in memory with the number of videos. We concatenate the result of each level into a feature vector $f \in \mathbb{R}^{LF}$, along with auxiliary inputs such as viewing direction, and pass the resulting vector into an MLP to obtain per-branch outputs.

\textbf{Static branch.} We generate RGB images by combining the outputs of our three branches. The static branch maps the feature vector obtained from the hash table into a view-dependent color $\textbf{c}_s$ and a view-independent density $\sigma_s$. To model lighting variations which could be dramatic across videos but smooth {\em within} a video, we condition on a latent embedding computed as a product of a video-specific matrix $A_{vid}$ and a fourier-encoded time index $\mathcal{F}(t)$ (as in~\cite{yang2022banmo}):
\begin{align}
    &\sigma_s(\textbf{x}) \in \mathbb{R} \\
    &\textbf{c}_s(\textbf{x}, \textbf{d}, A_{vid}\mathcal{F}(t)) \in \mathbb{R}^3.
\end{align}

\textbf{Dynamic branch.} While the static branch assumes the density $\sigma_s$ is static,
the dynamic branch allows both the density $\sigma_d$ and color $\textbf{c}_d$ to depend on time (and video). We therefore omit the latent code when computing the dynamic radiance. Because we find shadows to play a crucial role in the appearance of urban scenes (Fig.~\ref{fig:shadows}), we explicitly model a \textit{shadow field} of scalar values $\rho_d \in [0, 1]$, used to scale down the static color $\textbf{c}_s$ (as done in \cite{wu2022d}):
\begin{figure}
\vspace*{-2mm}
     \centering
     \begin{subfigure}[b]{0.7\linewidth}
         \centering
         \includegraphics[width=\textwidth]{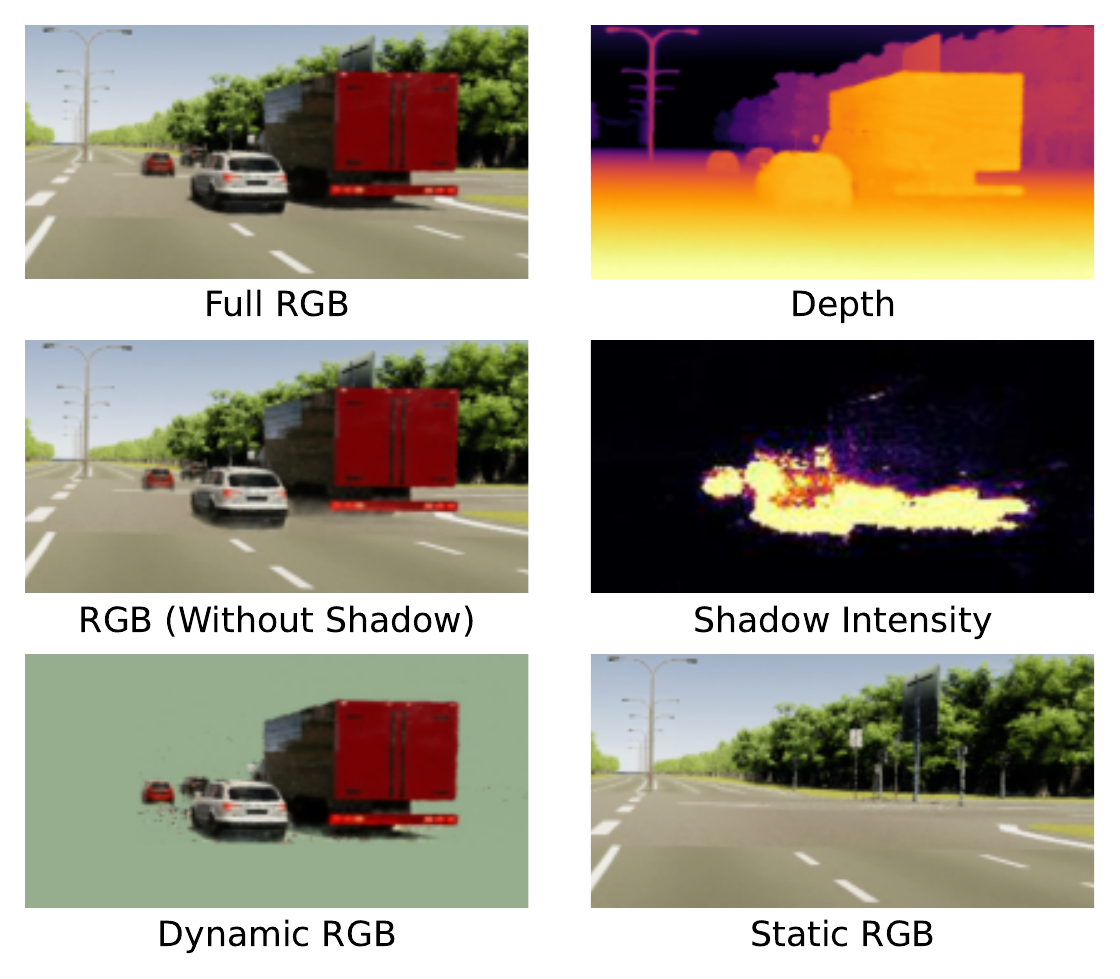}
         \caption*{\textbf{(a) Shadow Field}}
     \end{subfigure}
     \hfill
     \begin{subfigure}[b]{0.7\linewidth}
         \centering
         \includegraphics[width=\textwidth]{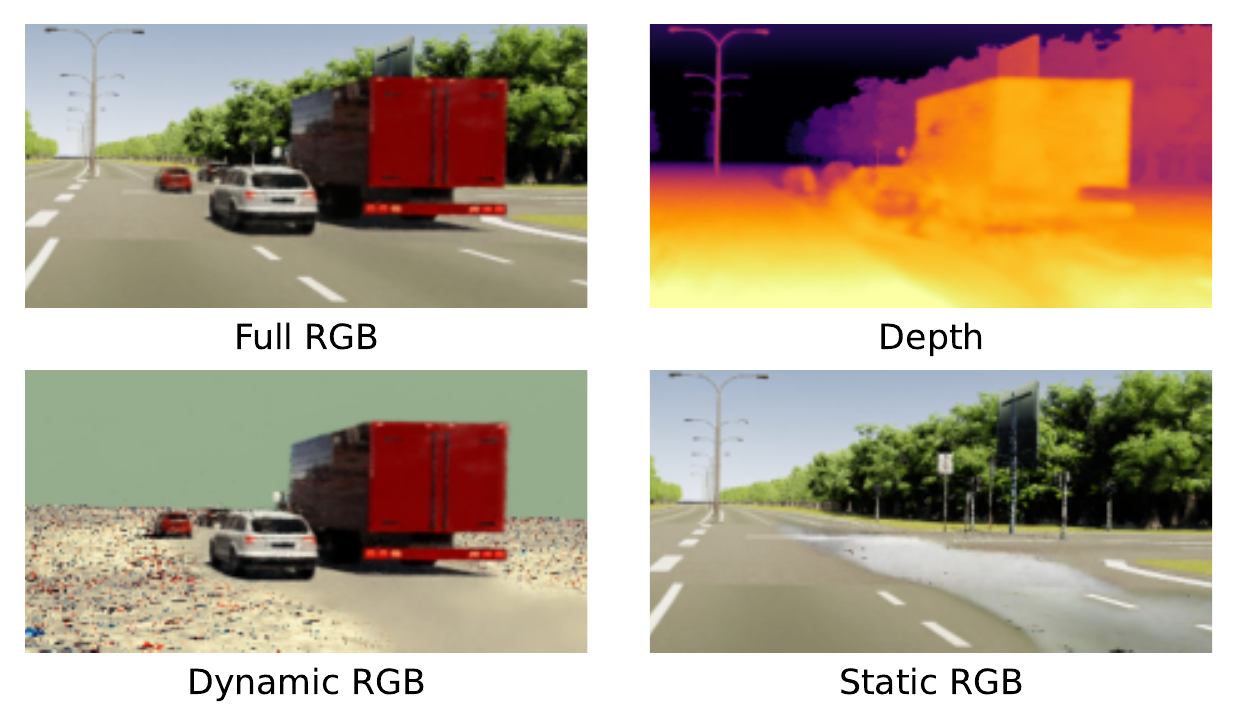}
         \caption*{\textbf{(b) No Shadow Field}}
     \end{subfigure}
\vspace*{-2mm}
\caption{{\bf Shadows.} We learn an explicit shadow field (\textbf{a}) as a pointwise reduction on static color, enabling better depth reconstruction and static/dynamic factorization than without (\textbf{b}).}
\vspace*{-4mm}
\label{fig:shadows}
\end{figure}
\begin{align}
    &\sigma_d(\textbf{x}, \textbf{t}, \textbf{vid}) \in \mathbb{R} \\
    &\rho_d(\textbf{x}, \textbf{t}, \textbf{vid}) \in [0, 1]   \\
    &\textbf{c}_d(\textbf{x}, \textbf{t}, \textbf{vid}, \textbf{d}) \in \mathbb{R}^3
\end{align}

\textbf{Far-field branch.} Because the sky requires reasoning about far-field radiance and because it can change dramatically across videos, we model far-field radiance with an environment map $\textbf{c}_e(\textbf{d}, \textbf{vid}) \in \mathbb{R}^3$ that depends on viewing direction ${\bf d}$~\cite{rematas2022urf, hao2021GANcraft} and a video id ${\bf vid}$.

{\bf Rendering.} We derive a single density and radiance value for any position by computing the weighted sum of the static and dynamic components, combined with the pointwise shadow reduction:
\begin{align}
    \sigma(\textbf{x}, \textbf{t}, \textbf{vid}) &= \sigma_s(\textbf{x}) + \sigma_d(\textbf{x}, \textbf{t}, \textbf{vid})  \\
    \textbf{c}(\textbf{x}, \textbf{t}, \textbf{vid}, \textbf{d}) &= \dfrac{\sigma_s}{\sigma}(1 - \rho_d)\textbf{c}_s(\textbf{x}, \textbf{d}, A_{vid}\mathcal{F}(t)) \nonumber\\
    &+ \dfrac{\sigma_d}{\sigma}\textbf{c}_d(\textbf{x}, \textbf{t}, \textbf{vid}, \textbf{d})
    \label{eq:composite}
\end{align}

We then calculate the color $\hat{C}$ for a camera ray $\textbf{r}$ with direction $\textbf{d}$ at a given frame $\textbf{t}$ and video $\textbf{vid}$ by accumulating the transmittance along sampled points $\textbf{r}(t)$ along the ray, forcing the ray to intersect the far-field environment map if it does not hit geometry within the foreground:
\begin{align}
    &\hat{C}(\textbf{r}, \textbf{t}, \textbf{vid}) = \int_0^{+\infty} T(t) \sigma(\textbf{r}(t), \textbf{t}, \textbf{vid}) \textbf{c}(\textbf{r}(t), \textbf{t}, \textbf{vid}, \textbf{d}) dt \nonumber \label{eq:volrend} \\
    & \quad\quad\quad\quad\quad\ + T(+\infty) \textbf{c}_e(\textbf{d}, \textbf{vid}),\\
    &\mathrm{where}~ T(t) = \exp{\left( - \int_0^t \sigma(\textbf{r}(s), \textbf{t}, \textbf{vid})ds \right)}.
\end{align}

\textbf{Feature distillation.} We build semantic awareness into \method\ to enable the open-world tasks described in Sec.~\ref{sec:city-scale-eval}. Similar to recent work~\cite{kobayashi2022distilledfeaturefields, tschernezki22neural}, we distill the outputs of a self-supervised 2D feature extractor, namely DINO~\cite{caron2021emerging}, as a teacher model into our network. For a feature extractor that transforms an image into a dense $\mathbb{R}^{H \times W \times C}$ feature grid, we add a $C$-dimensional output head to each of our branches:
\begin{align}
    &\Phi_s(\textbf{x}) \in \mathbb{R}^C \\
    &\Phi_d(\textbf{x}, \textbf{t}, \textbf{vid}) \in \mathbb{R}^C \\
    &\Phi_e(\textbf{d}, \textbf{vid}) \in \mathbb{R}^C,
\end{align}

which are combined into a single value $\Phi$ at any 3D location and rendered into $\hat{F}(\textbf{r})$ per camera ray, following the equations for color (\ref{eq:composite}, \ref{eq:volrend}).

\textbf{Scene flow.} We train our model to predict 3D scene flow and model scene dynamics. Inspired by previous work~\cite{li2020neural, du2021nerflow, Gao-ICCV-DynNeRF}, we augment our dynamic branch to predict forward and backward 3D scene flow vectors $s_{t' \in [-1, 1]}(\textbf{x}, \textbf{t}, \textbf{vid}) \in \mathbb{R}^3$. We make use of these vectors to enforce consistency between observed time steps through multiple loss terms (Sec.~\ref{sec:losses}), which we find crucial to generating plausible renderings at novel time steps (Table~\ref{table:diagnostics}).

\textbf{Spatial partitioning.} We scale our representation to arbitrarily large environments by decomposing the scene into individually trained models~\cite{Turki_2022_CVPR, tancik2022blocknerf}, each with its own static, dynamic, and far-field branch. Intuitively, the reconstruction for neighborhood X can be done largely independantly of the reconstruction in neighborhood Y, provided one can assign the relevant input data to each reconstruction. To do so, we follow the approach of Mega-NeRF~\cite{Turki_2022_CVPR} and split the scene into $K$ spatial cells with centroids $k \in \mathbb{R}^3$. Crucially, we generate separate training datasets for each spatial cell by making use of {\em visibility} reasoning~\cite{funkhouser1992management}. Mega-NeRF includes only those datapoints whose associated camera rays intersect the spatial cell. However, this may still include datapoints that are not visible due to an intervening occluder (e.g., a particular camera in neighborhood X can be {\em pointed} at neighborhood Y, but may not see anything there due to occluding buildings). To remedy this, we make use of depth measurements to prune irrelevant pixel rays that do not terminate within the spatial cell of interest (making use of nearest-neighbor interpolation to impute depth for pixels without a LiDAR depth measurement). This further reduces the size of each trainset by 2x relative to Mega-NeRF. Finally, given such separate reconstructions, one can still produce a globally consistent rendering by querying the appropriate spatial cell when sampling points along new-view rays (as in~\cite{Turki_2022_CVPR}).

\begin{figure}
\centering
\includegraphics[width=0.8\linewidth, clip = true, trim = 0mm 0mm 0mm 0mm]{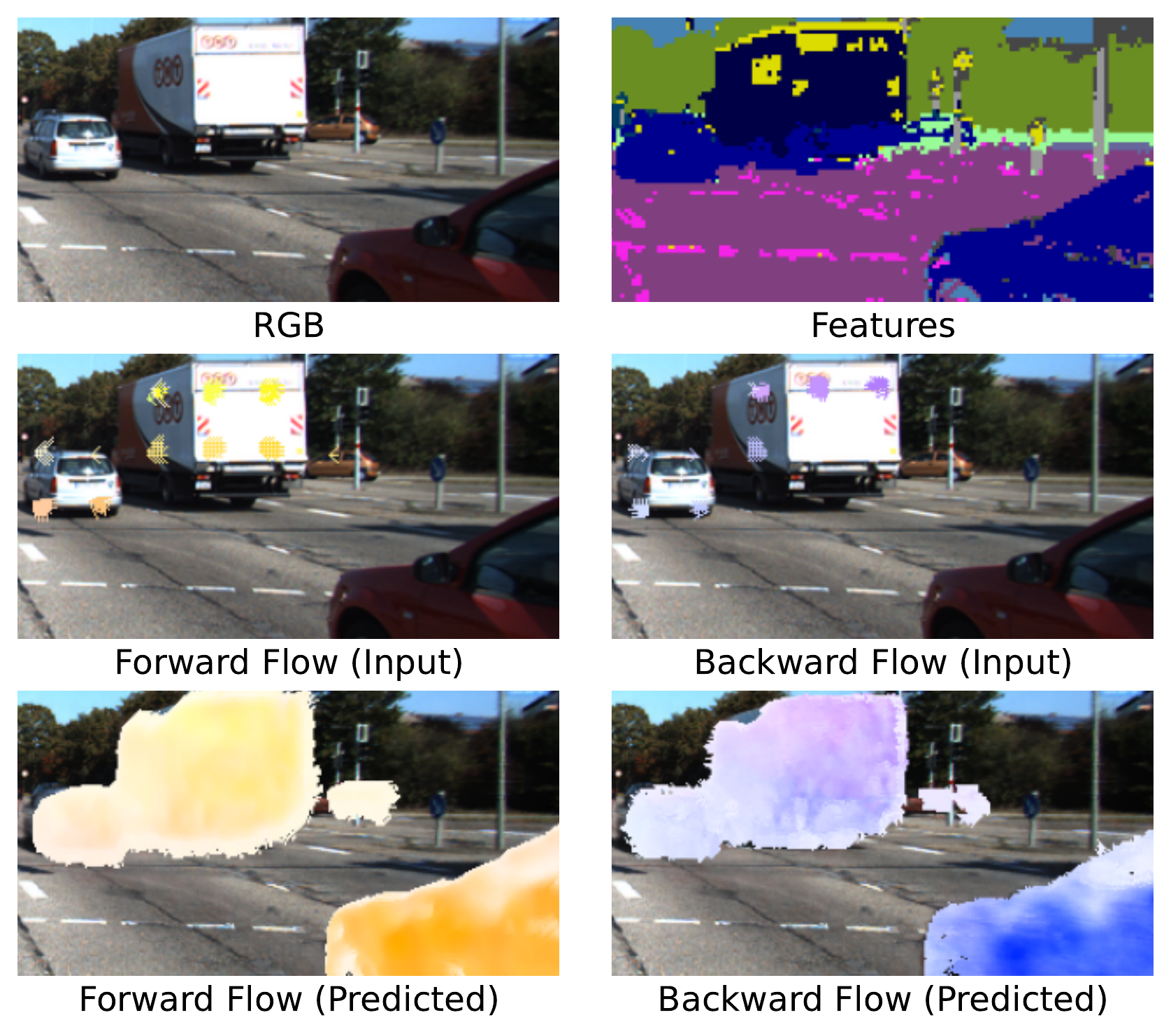}
\vspace*{-2mm}
  \caption{{\bf Scene Flow.} We minimize the photometric and feature-metric loss of warped renderings relative to ground truth inputs \textbf{(top)}. We use 2D optical flow from off-the-shelf estimators or sparse correspondences computed directly from 2D DINO features~\cite{amir2021deep} \textbf{(middle)} to supervise our flow predictions \textbf{(bottom)}.}
\label{fig:flow-overview}
\vspace*{-5mm}
\end{figure}

\subsection{Optimization}
\label{sec:losses}

We jointly optimize all three of our model branches along with the per-video weight matrices $A_{vid}$ by sampling random batches of rays across our $N$ input videos and minimizing the following loss:
\begin{equation}
\begin{aligned}
    \mathcal{L} &= {\underbrace {\Big(\mathcal{L}_c + \lambda_f \mathcal{L}_f + \lambda_d \mathcal{L}_d + \lambda_o \mathcal{L}_o \Big)}_\text{reconstruction losses}}  + {\underbrace {\Big(\mathcal{L}^w_c + \lambda_f \mathcal{L}^w_f \Big)}_\text{warping losses}} \\
    & \lambda_{flo}{\underbrace {\Big(\mathcal{L}_{cyc} + \mathcal{L}_{sm} + \mathcal{L}_{slo} \Big)}_\text{flow losses}} + {\underbrace {\Big(\lambda_e\mathcal{L}_{e} + \lambda_d\mathcal{L}_{d}\Big)}_\text{static-dynamic factorization}} + \lambda_{\rho}\mathcal{L}_{\rho}.
\end{aligned}
\end{equation}

\textbf{Reconstruction losses.} We minimize the L2 photometric loss $\mathcal{L}_{c}(\mathbf{r}) = \big\lVert{C(\textbf{r}) - \hat{C}(\textbf{r})}\big\rVert^2$ as in the original NeRF equation~\cite{mildenhall2020nerf}. We similarly minimize the L1 difference  $\mathcal{L}_{f}(\textbf{r}) = \big\lVert{F(\textbf{r}) - \hat{F}(\textbf{r})}\big\rVert_1$ between the feature outputs of the teacher model and that of our network.

To make use of our depth measurements, we project the LiDAR sweeps onto the camera plane and compare the expected depth $\hat{D}(r)$ with the measurement $D(\textbf{r})$~\cite{kangle2021dsnerf, rematas2022urf}:
\begin{align}
&\mathcal{L}_{d}(\mathbf{r}) = \big\lVert{D(\textbf{r}) - \hat{D}(\textbf{r})}\big\rVert^2 \\
&\mathrm{where}~ \hat{D}(\textbf{r}) = \int_0^{+\infty} T(s) \sigma(\textbf{r}(s))ds
\end{align}

{\bf Flow.} 
We supervise our 3D scene flow predictions based on 2D optical flow (Sec.~\ref{sec:experimental-setup}).  We generate a 2D displacement vector for each camera ray by first predicting its position in 3D space as the weighted sum of the scene flow neighbors along the ray:
\begin{align}
    \hat{X}_{t'}(\textbf{r}) = \int_0^{+\infty} T(t) \sigma(r(t))(r(t) + s_{t'}(\textbf{r}(t))) dt 
\end{align}
which we then ``render" into 2D using the camera matrix of the neighboring frame index. We minimize its distance from the observed optical flow via $\mathcal{L}_{o}(\textbf{r}) = \sum_{t' \in [-1, 1]}\big\lVert{X(\textbf{o}) - \hat{X}_{t'}(\textbf{r})}\big\rVert_1$. We anneal $\lambda_o$ over time as these estimates are noisy.

\textbf{3D warping.} 
The above loss ensures that rendered 3D flow will be consistent with the observed 2D flow. We also found it useful to enforce 3D color (and feature) constancy; i.e., colors remain constant even when moving. To do so, we use the predicted forward and backward 3D flow $s_{\textbf{t}+1}$ and $s_{\textbf{t}-1}$ to {\em advect} each sample along the ray into the next/previous frame:
\begin{align}
    &\sigma^w_{t'}(\textbf{x} + s_{t'}, \textbf{t} + t', \textbf{vid}) \in \mathbb{R} \\
    &\textbf{c}^w_{t'}(\textbf{x} + s_{t'}, \textbf{t} + t', \textbf{vid}, \textbf{d}) \in \mathbb{R}^3 \\
    &\Phi^w_{t'}(\textbf{x} + s_{t'}, \textbf{t} + t', \textbf{vid}) \in \mathbb{R}^C
\end{align}

The warped radiance $\textbf{c}^w$ and density $\sigma^w$ are rendered into warped color $\hat{C}^w(\textbf{r})$ and feature $\hat{F}^w(\textbf{r})$ (\ref{eq:composite}, \ref{eq:volrend}). We add a loss to ensure that the warped color (and feature) match the ground-truth input for the current frame, similar to ~\cite{li2020neural, Gao-ICCV-DynNeRF}. As in NSFF~\cite{li2020neural}, we found it important to downweight this loss in ambiguous regions that may contain occlusions. However, instead of learning explicit occlusion weights, we take inspiration from Kwea's method~\cite{nsff_pl} and use the difference between the dynamic geometry and the warped dynamic geometry to downweight the loss:
\begin{align}
    &w_{t'}(\textbf{x}, \textbf{t}, \textbf{vid}) = \left\lvert \frac{\sigma_d}{\sigma} - \frac{\sigma^w_{t'}}{\sigma} \right\rvert \\
    &\hat{W}_{t'}(\textbf{r}) = \int_0^{+\infty} T(t) \sigma(r(t))w_{t'}(r(t)) dt \label{eq:occlusion-weights}
\end{align}

resulting in the following warping loss terms:
\begin{align}
    &\mathcal{L}^w_{c}(\mathbf{r}) = \sum_{t' \in [-1, 1]} (1 - {W}_{t'})(\textbf{r}))\big\lVert{C(\textbf{r}) - \hat{C}^w_{t'}(\textbf{r})}\big\rVert^2 \label{eq:warped-color} \\
    &\mathcal{L}^w_{f}(\textbf{r}) = \sum_{t' \in [-1, 1]} (1 - {W}_{t'})(\textbf{r}) \big\lVert{F(\textbf{r}) - \hat{F}^w_{t'}(\textbf{r})}\big\rVert_1  \label{eq:warped-feature}
\end{align}

\textbf{Flow regularization.} As in prior work~\cite{li2020neural, Gao-ICCV-DynNeRF} we use a 3D scene flow cycle term to encourage consistency between forward and backward scene flow predictions, down-weighing the loss in areas ambiguous due to occlusions:
\begin{equation}
    \footnotesize{
    \mathcal{L}_{cyc}(\textbf{r}) = \sum_{t' \in [-1, 1]} \sum_\textbf{x} w_{t'}(\textbf{x}, \textbf{t}) \big\lVert{s_{t'}(\textbf{x}, \textbf{t}) + s_{\textbf{t}}(\textbf{x} + s_{t'}, \textbf{t} - t')}\big\rVert_1,
    }
\end{equation}

with $\textbf{vid}$ omitted for brevity. We also encourage spatial and temporal smoothness through $\mathcal{L}_{sm}(\textbf{r})$ as described in Sec.~\ref{sec:smoothness-priors}.
We finally regularize the magnitude of predicted scene flow vectors to encourage the scene to be static through $\mathcal{L}_{slo}(\textbf{r}) = \sum_{t' \in [\textbf{t} - 1, \textbf{t} + 1]} \sum_\textbf{x} \big\lVert{s_{t'}(\textbf{x}, \textbf{t})}\big\rVert_1$.

\textbf{Static-dynamic factorization.} As physically plausible solutions should have any point in space occupied by \textit{either} a static or dynamic object, we encourage the spatial ratio of static vs dynamic density to either be 0 or 1 through a skewed binary entropy loss that favors static explanations of the scene~\cite{wu2022d}:
\begin{align}
&\mathcal{L}_{e}(\mathbf{r}) = \int_0^{+\infty}  H\left(\dfrac{\sigma_d(\textbf{r}(t))}{\sigma_s(\textbf{r}(t))+\sigma_d(\textbf{r}(t))}^k\right) \: dt \\
&\mathrm{where}~ H(x) = -(x \cdot log(x) + (1-x) \cdot log(1-x)), \nonumber
\end{align}

and with $k$ set to 1.75, and further penalize the maximum dynamic ratio $\mathcal{L}_{d}(\mathbf{r}) = \max(\frac{\sigma_d(\textbf{r}(t))}{\sigma_s + \sigma_d})$ along each ray.

\textbf{Shadow loss.} We penalize the squared magnitude of the shadow ratio $\mathcal{L}_\rho (\mathbf{r}) =  \int_0^{+\infty} \rho_d(\mathbf{r}(t))^2 \: dt$ along each ray to prevent it from over-explaining dark regions~\cite{wu2022d}.

\begin{figure*}[t!]
\centering
\includegraphics[width=\textwidth, clip = true, trim = 0mm 2mm 0mm 0mm]{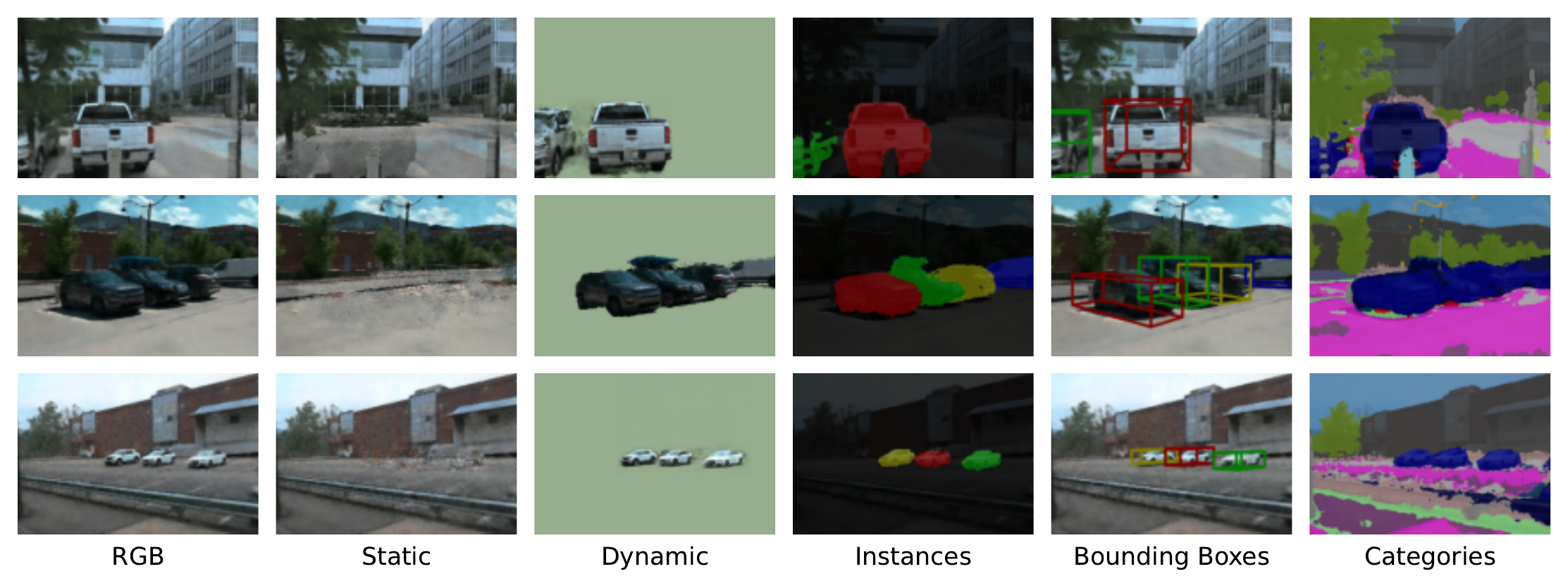}
\vspace*{-6mm}
   \caption{
   \textbf{City-1M.} We demonstrate \method's capabilities on multiple downstream tasks, including instance segmentation and 3D bounding box estimation without any labeled data (by just making use of geometric clustering). In the last column, we show category-level semantic classification by matching 3D (DINO) descriptors to a held-out video annotated with semantic labels. Please see text for more details.}
\label{fig:city-qualitative}
\vspace*{-4mm}
\end{figure*}

\section{Experiments}

We demonstrate \method's city-scale reconstruction capabilities by presenting quantitative results against baseline methods (Table~\ref{table:city-scale-reconstruction}). We also show initial qualitative results for a variety of downstream tasks (Sec.~\ref{sec:city-scale-eval}). Even though we focus on reconstructing dynamic scenes at city scale, to faciliate comparisons with prior work, we also show results on small-scale but highly-benchmarked datasets such as KITTI and Virtual KITTI 2 (Sec.~\ref{sec:kitti-benchmarks}). We evaluate the various components of our method in Sec.~\ref{sec:diagnostics}.

\subsection{Experimental Setup}
\label{sec:experimental-setup}

\textbf{2D feature extraction.} We use Amir et al's feature extractor implementation~\cite{amir2021deep} based on the dino\_vits8 model. We downsample our images to fit into GPU memory and then upsample with nearest neighbor interpolation. We L2-normalize the features at the 11th layer of the model and reduce the dimensionality to 64 through incremental PCA~\cite{incrementalpca}.

\textbf{Flow supervision.} We explored using an estimator trained on synthetic data~\cite{Teed2021RAFTRA} in addition to directly computing 2D correspondences from DINO itself~\cite{amir2021deep}. Although the correspondences are sparse (less than 5\% of pixels) and expensive to compute, we found its estimates more robust and use it for our experiments unless otherwise stated.

{\bf Training.} We train \method\ for 250,000 iterations with 4098 rays per batch and use a proposal sampling strategy similar to Mip-NeRF 360~\cite{barron2022mipnerf360} (Sec.~\ref{sec:proposal-sampling}). We use Adam~\cite{adam} with a learning rate of $5 \times 10^{-3}$ decaying to $5 \times 10^{-4}$.

{\bf Metrics.} We report quantitative results based on PSNR, SSIM~\cite{1284395}, and the AlexNet implementation of LPIPS~\cite{zhang2018perceptual}.

\subsection{City-Scale Reconstruction}
\label{sec:city-scale-eval}

\textbf{City-1M dataset.} We evaluate \method's large-scale reconstruction abilities on our collection of 1.28 million images across 1700 videos gathered across a 105 $km^2$ urban area using a vehicle-mounted platform with seven ring cameras and two LiDAR sensors. Due to the scale, we supervise optical flow with an off-the-shelf estimator trained on synthetic data~\cite{Teed2021RAFTRA} instead of DINO for efficiency.

\begin{table}[htbp!]
\centering
\resizebox{\linewidth}{!}{
\footnotesize
\setlength{\tabcolsep}{4pt}
\begin{tabular}{lcccc}
\toprule
 & Mega-NeRF~\cite{Turki_2022_CVPR} & Mega-NeRF-T & Mega-NeRF-A & \method \\ \midrule
PSNR $\uparrow$ & 16.42 & 16.46 & 16.70 & \textbf{21.67} \\
SSIM $\uparrow$ & 0.493 & 0.493 & 0.493 & \textbf{0.562}  \\
LPIPS $\downarrow$ & 0.879 & 0.877 & 0.850 & \textbf{0.554}  \\ \bottomrule
\end{tabular}
}
\caption{\textbf{City-scale view synthesis on City-1M.} \method\ outperforms all baselines by a wide margin.}
\label{table:city-scale-reconstruction}
\vspace*{-4mm}
\end{table}

\textbf{Baselines.} We compare \method\ to the official Mega-NeRF~\cite{Turki_2022_CVPR} implementation alongside two variants: Mega-NeRF-T which directly adds time as an input parameter to compute density and radiance, and Mega-NeRF-A which instead uses the latent embedding $A_{vid}\mathcal{F}(t)$ used by \method.

\textbf{Results.} We train both \method\ and the baselines using 48 cells and summarize our results in Table~\ref{table:city-scale-reconstruction}. \method\ outperforms all Mega-NeRF variants by a large margin. We provide qualitative results on view synthesis, static/dynamic factorization, unsupervised 3D instance segmentation and unsupervised 3D cuboid detection in Fig.~\ref{fig:city-qualitative}. We present additional qualititive tracking results in Fig.~\ref{fig:city-tracking}.

\textbf{Instance segmentation.} We derive the instance count as in prior work~\cite{see3d} by sampling dynamic density values $\sigma_d$, projecting those above a given threshold onto a discretized ground plane before applying connected component labeling. We apply k-means to obtain 3D centroids and volume render instance predictions as for semantic segmentation.

\textbf{3D cuboid detection.} After computing point-wise instance assignments in 3D, we derive oriented bounding boxes based on the PCA of the convex hull of points belonging to each instance~\cite{Open3D-boundingbox}.

\begin{figure*}[t!]
\centering
\includegraphics[width=\linewidth, clip = true, trim = 0mm 0mm 0mm 0mm]{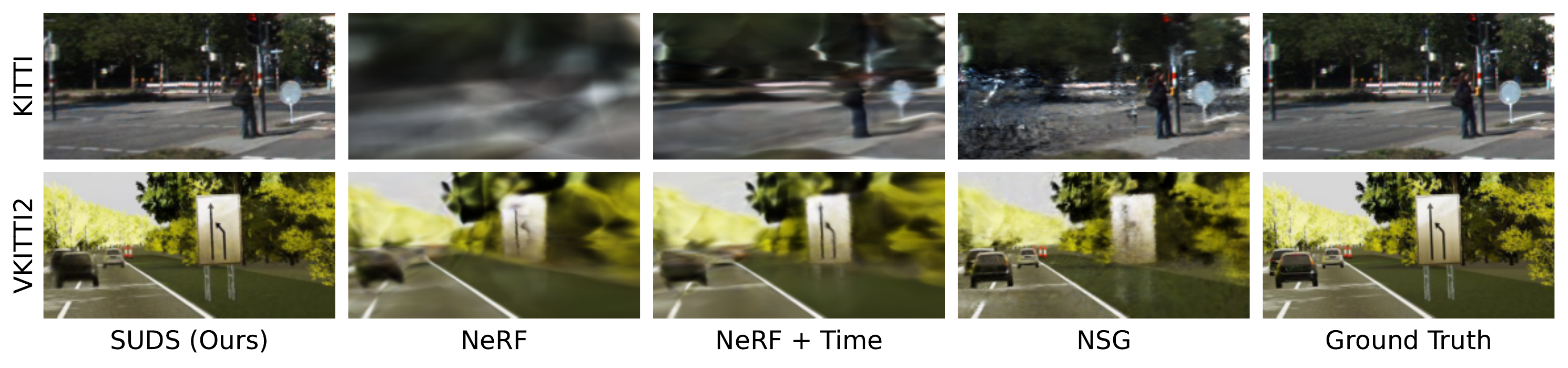}
\vspace*{-8mm}
  \caption{
  \textbf{KITTI and VKITTI2 view synthesis}. Prior work fails to represent the scene and NSG~\cite{Ost_2021_CVPR} renders ghosting artifacts near areas of movement. Our method forecasts plausible trajectories and generates higher-quality renderings.}
\label{fig:kitti-view-synthesis}
\end{figure*}

\begin{table*}[!htbp]
\centering
\footnotesize
\begin{tabular}{l@{\hspace{1em}}c@{\hspace{1em}}c@{\hspace{1em}}c@{\hspace{2em}}c@{\hspace{1em}}c@{\hspace{1em}}c@{\hspace{2em}}c@{\hspace{1em}}c@{\hspace{1em}}c@{\hspace{1em}}}
\toprule 
&\multicolumn{3}{c}{KITTI - 75\%} & \multicolumn{3}{c}{KITTI - 50\%} & \multicolumn{3}{c}{KITTI - 25\%} \\ \cmidrule(lr){2-4}\cmidrule(lr){5-7}\cmidrule(lr){8-10}
& $\uparrow$PSNR & $\uparrow$SSIM & $\downarrow$LPIPS & $\uparrow$PSNR & $\uparrow$SSIM & $\downarrow$LPIPS & $\uparrow$PSNR & $\uparrow$SSIM & $\downarrow$LPIPS \\ \midrule
NeRF~\cite{mildenhall2020nerf}\xspace & 18.56 & 0.557 & 0.554
& 19.12 & 0.587 & 0.497
& 18.61 & 0.570 & 0.510 \\
NeRF + Time\xspace & 21.01 & 0.612 & 0.492
& 21.34 & 0.635 & 0.448
& 19.55 & 0.586 & 0.505 \\
NSG~\cite{Ost_2021_CVPR}\xspace & 21.53 & 0.673 & 0.254
& 21.26 & 0.659 & 0.266
& 20.00	& 0.632 & 0.281 \\
\method\xspace  & \textbf{22.77} & \textbf{0.797} & \textbf{0.171}
& \textbf{23.12} & \textbf{0.821} & \textbf{0.135}
& \textbf{20.76} & \textbf{0.747}  & \textbf{0.198} \\
\midrule
&\multicolumn{3}{c}{VKITTI2 - 75\%} & \multicolumn{3}{c}{VKITTI2 - 50\%} & \multicolumn{3}{c}{VKITTI2 - 25\%} \\ \cmidrule(lr){2-4}\cmidrule(lr){5-7}\cmidrule(lr){8-10}
& $\uparrow$PSNR & $\uparrow$SSIM & $\downarrow$LPIPS & $\uparrow$PSNR & $\uparrow$SSIM & $\downarrow$LPIPS & $\uparrow$PSNR & $\uparrow$SSIM & $\downarrow$LPIPS \\ \midrule
NeRF~\cite{mildenhall2020nerf}\xspace & 18.67 & 0.548 & 0.634
& 18.58 & 0.544 & 0.635
& 18.17 & 0.537 & 0.644 \\
NeRF + Time\xspace & 19.03 & 0.574 & 0.587
& 18.90 & 0.565 & 0.610
& 18.04 & 0.545 & 0.626 \\
NSG~\cite{Ost_2021_CVPR}\xspace & 23.41 & 0.689 & 0.317
& 23.23 & 0.679 & 0.325
& 21.29 & 0.666 & 0.317 \\
\method\xspace & \textbf{23.87} & \textbf{0.846} & \textbf{0.150}
& \textbf{23.78} & \textbf{0.851} & \textbf{0.142}
& \textbf{22.18} & \textbf{0.829} & \textbf{0.160} \\
\bottomrule
\end{tabular}
\caption{{\bf Novel View Synthesis.} As the fraction of training views decreases, accuracy drops for all methods. However, \method\ consistently outperforms prior work, presumably due to more accurate representations learned by our diverse input signals (such as depth and flow).}
\vspace*{-4mm}
\label{table:novel-view-synthesis}
\end{table*}

\textbf{Semantic segmentation.} Note the above tasks of instance segmentation and 3D cuboid detection do not require any additional labels as they make use of geometric clustering. We now show that the representation learned by \method\ can also enable downstream semantic tasks, by making use of a small number of 2D segmentation labels provided on a held-out video sequence. We compute the average 2D DINO descriptor for each semantic class from the held out frames and derive 3D semantic labels for all reconstructions by matching each 3D descriptor to the closest class centroid. This allows to produce 3D semantic label fields that can then be rendered in 2D as shown in Fig.~\ref{fig:city-qualitative}.

\subsection{KITTI Benchmarks}
\label{sec:kitti-benchmarks}

\textbf{Baselines.} We compare \method\ to SRN~\cite{sitzmann2019srns}, the original NeRF implementation~\cite{mildenhall2020nerf}, a variant of NeRF taking time as an additional input, NSG~\cite{Ost_2021_CVPR}, and PNF~\cite{KunduCVPR2022PNF}. Both NSG and PNF are trained and evaluated using ground truth object bounding box and category-level annotations.

\textbf{Image reconstruction.} We compare \method's reconstruction capabilities using the same KITTI~\cite{Geiger2012CVPR} subsequences and experimental setup as prior work~\cite{Ost_2021_CVPR, KunduCVPR2022PNF}. We present results in Table~\ref{table:image-reconstruction}. As PNF's implementation is not publicly available, we rely on their reported numbers. \method\ surpasses the state-of-the-art in PSNR and SSIM.

\begin{table}
\centering
\resizebox{\linewidth}{!}{
\scriptsize
\setlength{\tabcolsep}{3.5pt}
\begin{tabular}{lcccccc}
\toprule
& SRN~\cite{sitzmann2019srns} & NeRF~\cite{mildenhall2020nerf}  & NeRF + Time & NSG~\cite{Ost_2021_CVPR}   & PNF~\cite{KunduCVPR2022PNF} & Ours           \\ \midrule
PSNR $\uparrow$ & 18.83 & 23.34 & 24.18 & 26.66 & 27.48 & \textbf{28.31} \\
SSIM $\uparrow$ & 0.590 & 0.662 & 0.677 & 0.806 & 0.870 & \textbf{0.876}  \\\bottomrule
\end{tabular}
}
\vspace*{-2mm}
\caption{ 
\textbf{KITTI image reconstruction.} We outperform past work on image reconstruction accuracy, following their experimental protocol and self-reported accuracies~\cite{Ost_2021_CVPR, KunduCVPR2022PNF}.}
\label{table:image-reconstruction}
\end{table}

\textbf{Novel view synthesis.} We demonstrate \method's capabilities to generate plausible renderings at time steps unseen during training. As NSG does not handle scenes with ego-motion, we use subsequences of KITTI and Virtual KITTI 2~\cite{gaidon2016virtual} with little camera movement. We evaluate the methods using different train/test splits, holding out every 4th time step, every other time step, and finally training with only one in every four time steps. We summarize our findings in Table~\ref{table:novel-view-synthesis} along with qualitative results in Fig.~\ref{fig:kitti-view-synthesis}. \method\ achieves the best results across all splits and metrics. Both NeRF variants fail to properly represent the scene, especially in dynamic areas. Although we provide NSG with the ground truth object poses at render time, it fails to learn a clean decomposition between objects and the background, especially as the number of training view decreases, and generates ghosting artifacts near areas of movement.

\begin{table}
\centering
\footnotesize
\begin{tabular}{lccc}
\toprule
 & $\uparrow$PSNR & $\uparrow$SSIM & $\downarrow$LPIPS  \\ \midrule
w/o Depth loss & 22.74 & 0.715 & 0.292 \\
w/o Optical flow loss & 22.18 & 0.708 & 0.302 \\
w/o Warping loss & 17.53 & 0.622 & 0.478  \\
w/o Appearance embedding & 22.54 & 0.704 & 0.296  \\
w/o Occlusion weights & 22.56 & 0.711 & 0.297  \\
w/o Separate branches & 19.73 & 0.570 & 0.475  \\
\cmidrule(lr){1-4}
Full Method & \textbf{22.95} & \textbf{0.718} & \textbf{0.289}  \\ \bottomrule
\end{tabular}
\caption{\textbf{Diagnostics.} Flow-based warping is the single-most important input, while depth is the least crucial input.}
\label{table:diagnostics}
\vspace*{-4mm}
\end{table}
\subsection{Diagnostics}
\label{sec:diagnostics}

We ablate the importance of major \method\ components by removing their respective loss terms along with occlusion weights, the latent embedding $A_{vid} \mathcal{F}(t)$ used to compute static color $\textbf{c}_s$, and separate model branches (Sec.~\ref{sec:ablations}). We run all approaches for 125,000 iterations across our datasets and summarize the results in Table~\ref{table:diagnostics}. Although all components help performance, flow-based warping is by far the single most important input. Interestingly, depth is the least crucial input, suggesting that \method\ can generalize to settings where depth measurements are not available.

\section{Conclusion}

We present a modular approach towards building dynamic neural representations at previously unexplored scale. Our multi-branch hash table structure enables us to disentangle and efficiently encode static geometry and transient objects across thousands of videos. \method\ makes use of unlabeled inputs to learn semantic awareness and scene flow, allowing it to perform several downstream tasks while surpassing state-of-the-art methods that rely on human labeling. Although we present a first attempt at building city-scale dynamic environments, many open challenges remain ahead of building truly photorealistic representations.

\section*{Acknowledgments}

This research was supported by the CMU Argo AI Center for Autonomous Vehicle Research.

\clearpage
{\small
\bibliographystyle{ieee_fullname}
\bibliography{main}
}

\clearpage
\appendix

\begin{figure*}[t!]
\centering
\includegraphics[width=\textwidth, clip = true, trim = 0mm 2mm 0mm 0mm]{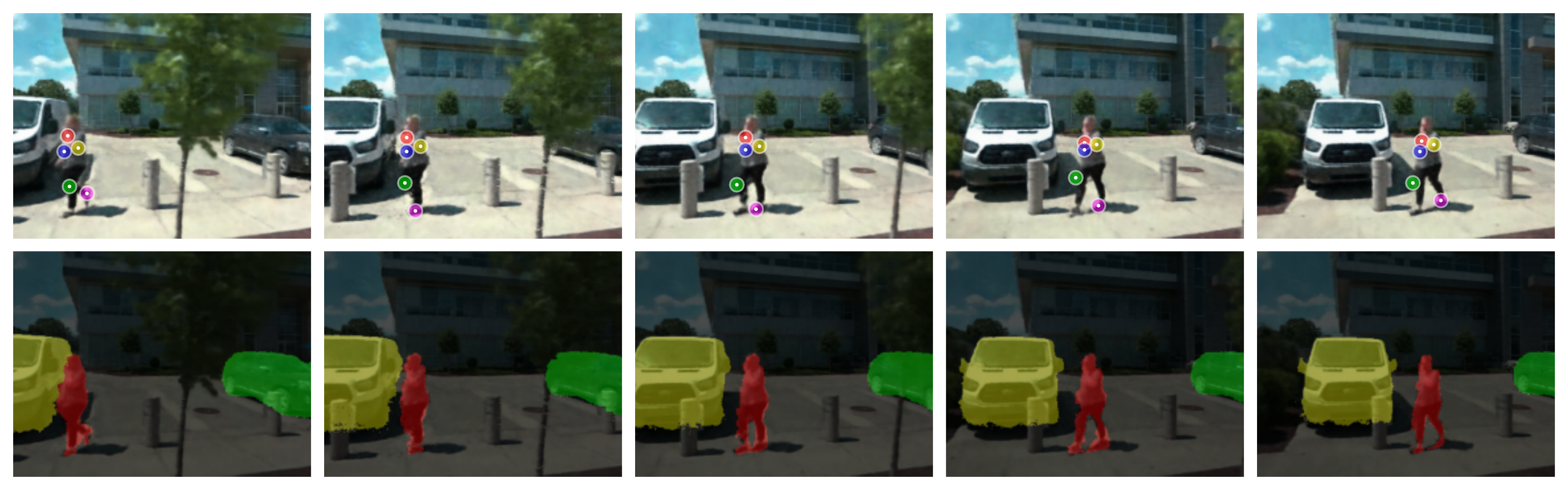}
   \caption{
   \textbf{Tracking.} We track keypoints (\textbf{above}) and instance masks (\textbf{below}) across several frames. As a 3D representation, \method\ can track correspondences through 2D occluders.}
\label{fig:city-tracking}
\end{figure*}

\section*{Supplemental Materials}

\section{Tracking}

We can compute mask and keypoint-level correspondences across frames after detecting instances (Sec.~\ref{sec:city-scale-eval}) by using Best-Buddies similarity~\cite{bestbuddies} on features $\Phi$ within or between instances. As a 3D representation, \method\ can track correspondences through 2D occluders. We show an example in Fig.~\ref{fig:city-tracking}.

\section{Proposal Sampling}
\label{sec:proposal-sampling}

We use a proposal sampling strategy similar to Mip-NeRF 360~\cite{barron2022mipnerf360} that first queries a lightweight occupancy proposal network at uniform intervals along each camera ray and then picks additional samples based on the initial samples. We model our proposal network with separate hash table-backed static and dynamic branches as in Sec.~\ref{sec:representation}. We train each branch of the proposal network with histogram loss~\cite{barron2022mipnerf360} using the weights of the respective branch of our main model and regularize the resulting sample distances and weights using distortion loss~\cite{barron2022mipnerf360}. We find that proposal sampling gives a 2-4x speedup.

\section{Smoothness Priors}
\label{sec:smoothness-priors}

We use the same spatial and temporal smoothness priors as NSFF~\cite{li2020neural} to regularize our scene flow. We specifically denote:
\begin{align}
    \mathcal{L}_{sm}(\textbf{r}) &= \sum_\textbf{x} \sum_{t' \in [-1, 1]}  e^{-2 \big\lVert{\textbf{x} - \textbf{x}'}\big\rVert_2}\big\lVert{s_{t'}(\textbf{x}, \textbf{t}) - s_{t'}(\textbf{x}', \textbf{t})}\big\rVert_1 \nonumber \\
    & +  \sum_\textbf{x} \big\lVert{s_{\textbf{t}-1}(\textbf{x}, \textbf{t}) + s_{\textbf{t}+1}(\textbf{x}, \textbf{t})}\big\rVert_1,
\end{align}
where $\textbf{x}$ and $\textbf{x}'$ indicate neighboring points along the camera ray $\textbf{r}$.

\section{Ablation Details}
\label{sec:ablations}

\textbf{w/o Depth loss.} We remove depth from the reconstruction loss term:
\begin{equation}
\mathcal{L}_{rec} = \mathcal{L}_c + \lambda_f \mathcal{L}_f + \lambda_o \mathcal{L}_o
\end{equation}

\textbf{w/o Optical flow loss.} We remove optical flow from the reconstruction loss term:
\begin{equation}
\mathcal{L}_{rec} = \mathcal{L}_c + \lambda_f \mathcal{L}_f + \lambda_d \mathcal{L}_d
\end{equation}

\textbf{w/o Warping loss.} We remove all warping and flow-related loss terms:
\begin{equation}
\mathcal{L} = {\underbrace {\Big(\mathcal{L}_c + \lambda_f \mathcal{L}_f + \lambda_d \mathcal{L}_d \Big)}_\text{reconstruction losses}} + {\underbrace {\Big(\lambda_e\mathcal{L}_{e} + \lambda_d\mathcal{L}_{d}\Big)}_\text{static-dynamic factorization}} + \lambda_{\rho}\mathcal{L}_{\rho}.
\end{equation}

\textbf{w/o Appearance embedding.} We compute static color without the latent embedding vector $A_{vid}\mathcal{F}(t)$: 

\begin{equation}
\textbf{c}_s(\textbf{x}, \textbf{d}) \in \mathbb{R}^3
\end{equation}

\textbf{w/o Occlusion weights.} We do not use occlusion weights (\ref{eq:occlusion-weights}) to downweight the warping loss terms (\ref{eq:warped-color}, \ref{eq:warped-feature}):

\begin{align}
    &\mathcal{L}^w_{c}(\mathbf{r}) = \sum_{t' \in [-1, 1]} \big\lVert{C(\textbf{r}) - \hat{C}^w_{t'}(\textbf{r})}\big\rVert^2 \\
    &\mathcal{L}^w_{f}(\textbf{r}) = \sum_{t' \in [-1, 1]}  \big\lVert{F(\textbf{r}) - \hat{F}^w_{t'}(\textbf{r})}\big\rVert_1 
\end{align}

\textbf{w/o Separate branches.} We generate all model outputs using a single time-dependent branch:

\begin{align}
    &\sigma(\textbf{x}, \textbf{t}, \textbf{vid}) \in \mathbb{R} \\
    &\textbf{c}(\textbf{x}, \textbf{t}, \textbf{vid}, \textbf{d}) \in \mathbb{R}^3 \\
    &\Phi(\textbf{x}, \textbf{t}, \textbf{vid}) \in \mathbb{R}^C \\
    &s_{t' \in [-1, 1]}(\textbf{x}, \textbf{t}, \textbf{vid}) \in \mathbb{R}^3
\end{align}

We accordingly remove factorization-related loss terms:

\begin{equation}
\begin{aligned}
    \mathcal{L} &= {\underbrace {\Big(\mathcal{L}_c + \lambda_f \mathcal{L}_f + \lambda_d \mathcal{L}_d + \lambda_o \mathcal{L}_o \Big)}_\text{reconstruction losses}}  + {\underbrace {\Big(\mathcal{L}^w_c + \lambda_f \mathcal{L}^w_f \Big)}_\text{warping losses}} \\
    & \lambda_{flo}{\underbrace {\Big(\mathcal{L}_{cyc} + \mathcal{L}_{sm} + \mathcal{L}_{slo} \Big)}_\text{flow losses}}
\end{aligned}
\end{equation}

\section{Additional Training Details}

We divide City-1M into 48 cells using camera-based k-means clustering. Each cell covers 2.9 $km^2$ and 32k frames across 98 videos on average. We evaluate the effect of geographic coverage and number of frames/videos on cell quality in Table~\ref{tab:city-diagnostics}. We train with 1 A100 (40 GB) GPU per cell for 2 days (same for each KITTI scene). We can fit all cells on a single A100 at inference time.

\begin{table*}[htbp!]
\centering
\footnotesize
\subcaptionbox*{\textbf{Images}}{
\begin{tabular}{lcccc}
\toprule
& $\leq$ 15k & 15-30k & 30-45k &  $\geq$ 45k   \\ \midrule
$\uparrow$PSNR     & 22.86 & 21.99 & 21.35 & 20.75  \\
$\uparrow$SSIM     & 0.583 & 0.569 & 0.557 & 0.538  \\
$\downarrow$LPIPS  & 0.516 & 0.545 & 0.564 & 0.578  \\ \bottomrule
\end{tabular}
}
\subcaptionbox*{\textbf{Videos}}{
\begin{tabular}{lcccc}
\toprule
& $\leq$ 60 & 60-90 & 90-120 &  $\geq$ 120   \\ \midrule
$\uparrow$PSNR     & 22.47 & 21.72 & 21.68 & 21.11 \\
$\uparrow$SSIM     & 0.587 & 0.556 & 0.559 & 0.555  \\
$\downarrow$LPIPS  & 0.526 & 0.557 & 0.557 & 0.565  \\ \bottomrule
\end{tabular}
}
\par\medskip
\subcaptionbox*{\textbf{Area}}{
\begin{tabular}{lcccc}
\toprule
& $\leq$ 2 $km^2$ & 2-3 $km^2$ & 3-4 $km^2$ &  $\geq$ 4 $km^2$   \\ \midrule
$\uparrow$PSNR     & 22.73 & 21.47 & 21.53 & 22.18  \\
$\uparrow$SSIM     & 0.609 & 0.556 & 0.561 & 0.557  \\
$\downarrow$LPIPS  & 0.512 & 0.564 & 0.555 & 0.536  \\ \bottomrule
\end{tabular}
}
\vspace*{-2mm}
\caption{\textbf{City-1M scaling.} We evaluate the effect of geographic coverage and the number of images and videos on cell quality. Although performance degrades sublinearly across all metrics, image and video counts have the largest impact.}
\vspace*{-6mm}
\label{tab:city-diagnostics}
\end{table*}

\section{Assets}

\textbf{City-1M.} Our dataset is constructed from street-level videos collected across a vehicle fleet with seven ring cameras that collect 2048x1550 resolution images at 20 Hz with a combined 360° field of view. Both VLP-32C LiDAR sensors are synchronized with the cameras and produce point clouds with 100,000 points at 10 Hz on average. We localize camera poses using a combination of GPS-based and sensor-based methods.

\textbf{Third-party assets.} We primarily base the \method\ implementation on Nerfstudio~\cite{nerfstudio} and tiny-cuda-nn~\cite{tiny-cuda-nn} along with various utilities from OpenCV~\cite{opencv_library}, Scikit~\cite{sklearn_api}, and Amir et al's feature extractor implementation~\cite{amir2021deep}, all of which are freely available for noncommercial use. KITTI~\cite{Geiger2012CVPR} is similarly available under an Apache license, whereas VKITTI2~\cite{gaidon2016virtual} uses the noncommercial CC BY-NC-SA 3.0 license.

\section{Limitations}

\textbf{Video boundaries.} Although our global representation of static geometry is consistent across all videos used for reconstruction, all dynamic objects are video-specific. Put otherwise, our method does not allow us to extrapolate the movement of objects outside of the boundaries of videos from which they were captured, nor does it provide a straightforward way of rendering dynamic visuals at boundaries where camera rays intersect regions with training data originating from disjoint video sequences.

\textbf{Camera accuracy.} Accurate camera extrinsics and intrinsics are arguably the largest contributors to high NeRF rendering quality. Although multiple efforts~\cite{lin2021barf, wang2021nerfmm, SCNeRF2021, meng2021gnerf, 10.1007/978-3-031-19827-4_16} attempt to jointly optimize camera parameters during NeRF optimization, we found the results lacking relative to using offline structure-from-motion based approaches as a preprocessing step.

\textbf{Flow quality.} Although our method tolerates some degree of noisiness in the supervisory optical flow input, high-quality flow still has a measurable impact on model performance (and completely incorrect supervision degrades quality). We also assume that flow is linear between observed timestamps to simplify our scene flow representation.

\textbf{Resources.} Modeling city scale requires a large amount of dataset preprocessing, including, but not limited to: extracting DINO features, computing optical flow, deriving normalized coordinate bounds, and storing randomized batches of training data to disk. Collectively, our intermediate representation required more than 20TB of storage even after compression.

\textbf{Shadows.} \method\ attempts to disentangle shadows underneath transient objects. However, if a shadow is present in all observations for a given location (eg: a parking spot that is always occupied, even by different cars), \method\ may attribute the darkness to the static topology, as evidenced in several of our videos, even if the origin of the shadow is correctly assigned to the dynamic branch.

\textbf{Instance-level tasks.} Although we provide initial qualitative results on instance-level tasks as a first step towards true 3D segmentation backed by neural radiance field, \method is not competitive with conventional approaches.

\section{Societal Impact}

As \method\ attempts to model dynamic urban scenes with pedestrians and vehicles, our approach carries surveillance and privacy concerns related to the intentional or inadvertent capture or privacy-sensitive information such as human faces and vehicle license plate numbers. As we distill semantic knowledge into \method, we are able to (imperfectly) filter out either entire categories (people) or components (faces) at render time. However this information would still reside in the model itself. This could in turn be mitigated by preprocessing the input data used to train the model.

\end{document}